\documentclass[twoside]{article}

\usepackage[accepted]{aistats2020}
\usepackage[utf8]{inputenc} 
\usepackage[T1]{fontenc}    
\usepackage{url}            
\usepackage{booktabs}       
\usepackage{amsfonts}       
\usepackage{nicefrac}       
\usepackage{microtype}      
\usepackage{amsthm}
\usepackage{amsmath}
\usepackage{amssymb}
\usepackage{mathrsfs}
\usepackage{mathtools,amsfonts,stmaryrd}
\usepackage{graphicx}
\usepackage{caption}
\usepackage{subcaption}
\usepackage{multirow}
\usepackage{paralist}
\usepackage{enumitem}
\usepackage{algorithm}
\usepackage{algorithmic}
\usepackage{arydshln}

\usepackage{natbib}

\definecolor{mydarkblue}{rgb}{0,0.08,0.45}
\usepackage[colorlinks=true,
    linkcolor=mydarkblue,
    citecolor=mydarkblue,
    filecolor=mydarkblue,
    urlcolor=mydarkblue]{hyperref}

\newcommand{\R}{\mathbb{R}}
\newcommand{\Rd}{\mathbb{R}^d}

\newcommand{\PdR}{\mathscr{P}_2(\mathbb{R})}

\newcommand{\PdRd}{\mathscr{P}_2(\mathbb{R}^d)}
\newcommand{\N}{\mathbb{N}}
\newcommand{\W}{W}
\newcommand{\Lebesgue}[1]{\mathscr{L}^{#1}}
\newcommand{\dotp}[2]{\ensuremath{\langle #1 , #2\,\rangle}}
\newcommand{\range}[1]{\llbracket#1\rrbracket}
\newcommand{\barycentric}[1]{\overline{#1}}
\newcommand{\lip}[1]{\text{Lip}(#1)}
\newcommand{\one}[1]{{\mathbf{1}\{#1\}}}
\newcommand{\scvx}{\ell}
\newcommand{\opt}[1]{{{#1}_\star}}

\def\ba{\mathbf{a}}
\def\bb{\mathbf{b}}

\def\bx{\mathbf{x}}
\def\bw{\mathbf{w}}
\def\bu{\mathbf{u^+}}
\def\buminus{\mathbf{u^-}}
\def\bv{\mathbf{v}}

\DeclareMathOperator*{\argmin}{arg\,min}

\DeclareMathOperator{\supp}{supp}
\DeclareMathOperator{\proj}{proj}
\DeclareMathOperator{\sign}{sign}
\DeclareMathOperator{\continuouswrt}{\ll}
\DeclareMathOperator{\expectation}{\mathbb{E}}

\def\restriction#1#2{\mathchoice
              {\setbox1\hbox{${\displaystyle #1}_{\scriptstyle #2}$}
              \restrictionaux{#1}{#2}}
              {\setbox1\hbox{${\textstyle #1}_{\scriptstyle #2}$}
              \restrictionaux{#1}{#2}}
              {\setbox1\hbox{${\scriptstyle #1}_{\scriptscriptstyle #2}$}
              \restrictionaux{#1}{#2}}
              {\setbox1\hbox{${\scriptscriptstyle #1}_{\scriptscriptstyle #2}$}
              \restrictionaux{#1}{#2}}}
\def\restrictionaux#1#2{{#1\,\smash{\vrule height .8\ht1 depth .85\dp1}}_{\,#2}}

\newtheorem{theorem}{Theorem}
\newtheorem{proposition}{Proposition}
\newtheorem{definition}{Definition}

\newtheorem{remark}{Remark}

\begin{document}

%
\runningtitle{Regularity as Regularization: Smooth and Strongly Convex Brenier Potentials in Optimal Transport}

%

\twocolumn[
\aistatstitle{Regularity as Regularization:\\Smooth and Strongly Convex Brenier Potentials in Optimal Transport}

\aistatsauthor{ Fran\c{c}ois-Pierre Paty \And Alexandre d'Aspremont \And  Marco Cuturi }

\aistatsaddress{ CREST, ENSAE,\\
  Institut Polytechnique de Paris
  
  \And CNRS, ENS,\\
  PSL Research University
  
  \And Google Brain,\\
  CREST, ENSAE }
]

\begin{abstract}
    Estimating Wasserstein distances between two high-dimensional densities suffers from the curse of dimensionality: one needs an exponential (wrt dimension) number of samples to ensure that the distance between two empirical measures is comparable to the distance between the original densities. Therefore, optimal transport (OT) can only be used in machine learning if it is substantially regularized. On the other hand, one of the greatest achievements of the OT literature in recent years lies in regularity theory: \citet{caffarelli2000monotonicity} showed that the OT map between two well behaved measures is Lipschitz, or equivalently when considering 2-Wasserstein distances, that Brenier convex potentials (whose gradient yields an optimal map) are smooth. We propose in this work to draw inspiration from this theory and use regularity as a regularization tool. We give algorithms operating on two discrete measures that can recover nearly optimal transport maps with small distortion, or equivalently, nearly optimal Brenier potentials that are strongly convex and smooth. The problem boils down to solving alternatively a convex QCQP and a discrete OT problem, granting access to the values and gradients of the Brenier potential not only on sampled points, but also out of sample at the cost of solving a simpler QCQP for each evaluation. We propose algorithms to estimate and evaluate transport maps with desired regularity properties, benchmark their statistical performance, apply them to domain adaptation and visualize their action on a color transfer task.
\end{abstract}

\section{INTRODUCTION}



Optimal transport (OT) has found practical applications in areas as diverse as supervised machine learning~\citep{FrognerNIPS,abadeh2015distributionally,courty2016optimal}, graphics~\citep{2015-solomon-siggraph,2016-bonneel-barycoord}, generative models~\citep{WassersteinGAN,salimans2018improving}, NLP~\citep{grave2018unsupervised,alaux2018unsupervised}, biology~\citep{hashimoto2016learning,schiebinger2019optimal} or imaging~\citep{rabin2015convex,2016-Cuturi-siims}. OT theory is useful in these applications because it provides tools that can quantify the closeness between probability measures even when they do not have overlapping supports, and more generally because it defines tools to infer maps that can push-forward (or morph) one measure onto another. There is, however, an important difference between the OT definitions introduced in celebrated textbooks by~\citet{Villani03,Villani09} and~\citet{SantambrogioBook}, and those used in the works cited above. In all of these applications, some form of regularization is used to ensure that computations are not only tractable but also meaningful, in the sense that the naive implementation of linear programs to solve OT on discrete histograms/measures are not only too costly but also suffer from the curse of dimensionality~\citep{dudley1969speed, annurev-statistics-030718-104938}. Regularization, defined explicitly or implicitly as an approximation algorithm, is therefore crucial to ensure that OT is meaningful and can work at scale.

\textbf{Brenier Potentials and Regularity Theory.} In the OT literature, regularity has a different meaning, one that is usually associated with the properties of the optimal Monge map~\cite[\S9-10]{Villani09} pushing forward a measure $\mu$ onto $\nu$ with a small average cost. When that cost is the quadratic Euclidean distance, the Monge map is necessarily the gradient $\nabla f$ of a convex function $f$. This major result, known as~\citet{MR923203} theorem, states that the OT problem between $\mu$ and $\nu$ is solved as soon as there exists a convex function $f$ such that $\nabla f_\sharp \mu=\nu$. In that context, regularity in OT is usually understood as the property that the map $\nabla f$ is \textit{Lipschitz}, a seminal result due to~\citet{caffarelli2000monotonicity} who proved that the Brenier map can be guaranteed to be 1-Lipschitz when transporting a ``fatter than Gaussian'' measure $\mu\propto e^{V}\gamma_d$ towards a ``thinner than Gaussian'' measure $\nu\propto e^{-W}\gamma_d$ (here $\gamma_d$ is the Gaussian measure on $\R^d$, $\gamma_d\propto e^{-\|\cdot\|^2}$, and $V,W$ are two convex potentials). Equivalently, this result can be stated as the fact that the Monge map is the gradient of a $1$-smooth~\citet{MR923203} potential.

\textbf{Contributions.} Our goal in this work is to translate the idea that the OT map between sufficiently well-behaved distributions should be regular into an estimation procedure. Our contributions are:
\vspace{-.2cm}
\begin{enumerate}[leftmargin=15pt,itemsep=0pt]
    \item Given two probability measures $\mu,\nu\in\PdRd$, a $L$-smoooth and $\scvx$-strongly convex function $f$ such that $\nabla f_\sharp\mu =\nu$ may not always exist. We relax this equality and look instead for a smooth strongly convex function $f$ that minimizes the Wasserstein distance between $\nabla f_{\sharp}\mu$ and $\nu$. We call such potential nearest-Brenier because they provide the ``nearest'' way to transport $\mu$ to a measure like $\nu$ using a smooth and strongly convex potential.
    \item When $\mu,\nu$ are discrete probability measures, we show that nearest-Brenier potentials can be recovered as the solution of a QCQP/Wasserstein optimization problem. Our formulation builds upon recent advances in mathematical programming to quantify the worst-case performance of first order methods when used on smooth strongly convex functions~\citep{taylor2017smooth,drori2014performance}, yet results in simpler, convex problems.
    \item In the univariate case, we show that computing the nearest-Brenier potential is equivalent to solving a variant of the isotonic regression problem in which the map must be strongly increasing and Lipschitz. A projected gradient descent approach can be used to solve this problem efficiently.
    \item We exploit the solutions to both these optimization problems to extend the Brenier potential and Monge map at any point. We show this can be achieved by solving a QCQP for each new point.
    \item We implement and test these algorithms on various tasks, in which smooth strongly convex potentials improve the statistical stability of the estimation of Wasserstein distances, and illustrate them on color transfer and domain adaptation tasks.
\end{enumerate}
\section{REGULARITY IN OPTIMAL TRANSPORT}\label{sec:reginOT}

For $d \in \N$, we write $\range{d} = \{1, ..., d\}$ and $\Lebesgue{d}$ for the Lebesgue measure in $\Rd$. We write $\PdRd$ for the set of Borel probability measures with finite second-order moment.

\textbf{Wasserstein distances, Kantorovich and Monge Formulations.} For two probability measures $\mu, \nu \in \PdRd$, we write $\Pi(\mu, \nu)$ for the set of couplings
\begin{multline*}
        \Pi(\mu, \nu) = \{ \pi \in \mathscr{P}(\Rd \times \Rd) \textrm{ s.t.} \,\forall A, B\subset\Rd \text{ Borel},\\
        \pi(A\times \Rd)=\mu(A), \pi(\Rd \times B)=\nu(B) \},
\end{multline*}
and define their $2$-Wasserstein distance as the solution of the Kantorovich problem~\cite[\S6]{Villani09}:
\[
    \W_2(\mu, \nu) := \left( \inf_{\pi \in \Pi(\mu, \nu)} \int \|x-y\|_2^2 \,d\pi(x,y) \right)^{1/2}.
\]
For Borel sets $\mathcal{X}, \mathcal{Y} \subset \Rd$, Borel map $T : \mathcal{X} \to \mathcal{Y}$ and $\mu \in \mathscr{P}(\mathcal{X})$, we denote by $T_\sharp\mu \in \mathscr{P}(\mathcal{Y})$ the push-forward of $\mu$ under $T$, \emph{i.e.} the measure such that for any $A \subset \mathcal{Y}$, $T_\sharp\mu(A) = \mu\left( T^{-1}(A) \right)$.
The~\citet{Monge1781} formulation of OT consists in considering maps such that $T_\sharp\mu=\nu$, instead of couplings. Both formulations are equal when feasible maps exist, namely 
\[
    \W_2(\mu, \nu) = \left( \inf_{T: T_\sharp\mu=\nu} \int \|x-T(x)\|^2 \,d\mu(x) \right)^{1/2}.
\]

\textbf{Convexity and Transport: The Brenier Theorem.}
Let $\mu \in \PdRd$ and $f: \Rd \to \R$ convex and differentiable $\mu$-a.e. Then $\nabla f$, as a map from $\Rd$ to $\Rd$ is optimal for the Monge formulation of OT between the measures $\mu$ and $\nabla f_\sharp\mu$. The~\citet{MR923203} theorem shows that if $\mu= p \Lebesgue{d}$ ($\mu$ is absolutely continuous w.r.t. $\Lebesgue{d}$ with density $p$) and $\nu \in \PdRd$, there always exists a convex $f$ such that $\nabla f_\sharp\mu = \nu$, \emph{i.e.} there exists an optimal Monge map sending $\mu$ to $\nu$ that is the gradient of a convex function $f$. Such a convex function $f$ is called a Brenier potential between $\mu$ and $\nu$. If moreover $\nu = q \Lebesgue{d}$, that is $\nu$ has density $q$, a change of variable formula shows that $f$ should be solution to the Monge-Amp\`ere~\cite[Eq.12.4]{Villani09} equation $\det(\nabla^2 f) = \frac{p}{q \circ \nabla f}$. The study of the Monge-Amp\`ere equation is the key to obtain regularity results on $f$ and $\nabla f$, see the recent survey by~\citet{figalli2017monge}.

\textbf{Strong Convexity and Smoothness.}
We recall that a differentiable convex function $f$ is called $L$-smooth if its gradient function is $L$-Lipschitz, namely for all $x,y\in\R^d$ we have $\|\nabla f(x)-\nabla f(y)\|\leq L \|x-y\|$. It is called $\scvx$-strongly convex if $f - (\scvx/2)\|\cdot\|^2$ is convex. Given a partition $\mathcal{E}=(E_1,\dots,E_K)$ of $\Rd$, we will more generally say that $f$ is $\mathcal{E}$-locally $\scvx$-strongly convex and $L$-smooth if the inequality above only holds for pairs $(x,y)$ taken in the interior of any of the subsets $E_k$. We write $\mathcal{F}_{\scvx, L,\mathcal{E}}$ for the set of such functions.

\textbf{Regularity of OT maps.} Results on the regularity of the Brenier potential were first obtained by~\citet{caffarelli2000monotonicity}. For measures $\mu = e^{V}\gamma_d$ and $\nu = e^{-W}\gamma_d$, where $V,W$ are convex functions and $\gamma_d$ is the standard Gaussian measure on $\Rd$, Caffarelli's contraction theorem states that any Brenier potential $\opt{f}$ between $\mu$ and $\nu$ is $1$-smooth. More general results have been proposed by~\citet{FigalliPartial} who showed that local regularity holds in a general setting: loosely speaking, one can obtain ``local H\"older regularity by parts'' as soon as the measures have bounded densities and compact support.

\section{REGULARITY AS REGULARIZATION}\label{sec:regasreg}

Contrary to the viewpoint adopted in the OT literature~\citep{caffarelli1999problem,figalli2017monge}, we consider here regularity (smoothness) and curvature (strong convexity), as \textit{desiderata}, namely conditions that must be enforced when estimating OT, rather than properties that can be proved under suitable assumptions on $\mu,\nu$. Note that if a convex potential is $\scvx$-strongly convex and $L$-smooth, the map $\nabla f$ has distortion $\ell\|x-y\|\leq \|\nabla f(x)-\nabla f(y)\|\leq L\|x-y\|$. Therefore, setting $\scvx=L=1$ enforces that $\nabla f$ must be a translation, since $f$ must be convex. If one were to lift the assumption that $f$ is convex, one would recover the case where $\nabla f$ is an isometry, considered in~\citep{cohen1999earth,alt2000discrete,alaux2018unsupervised}. Note that this distortion also plays a role when estimating the Gromov-Wasserstein distance between general metric spaces~\citep{memoli-2011} and was notably used to enforce regularity when estimating the discrete OT problem~\citep{flamary2014optimal} in a Kantorovich setting. We enforce it here as a constraint in the space of convex functions.

\begin{figure}[!h]
    \centering
    \includegraphics[width=0.48\textwidth]{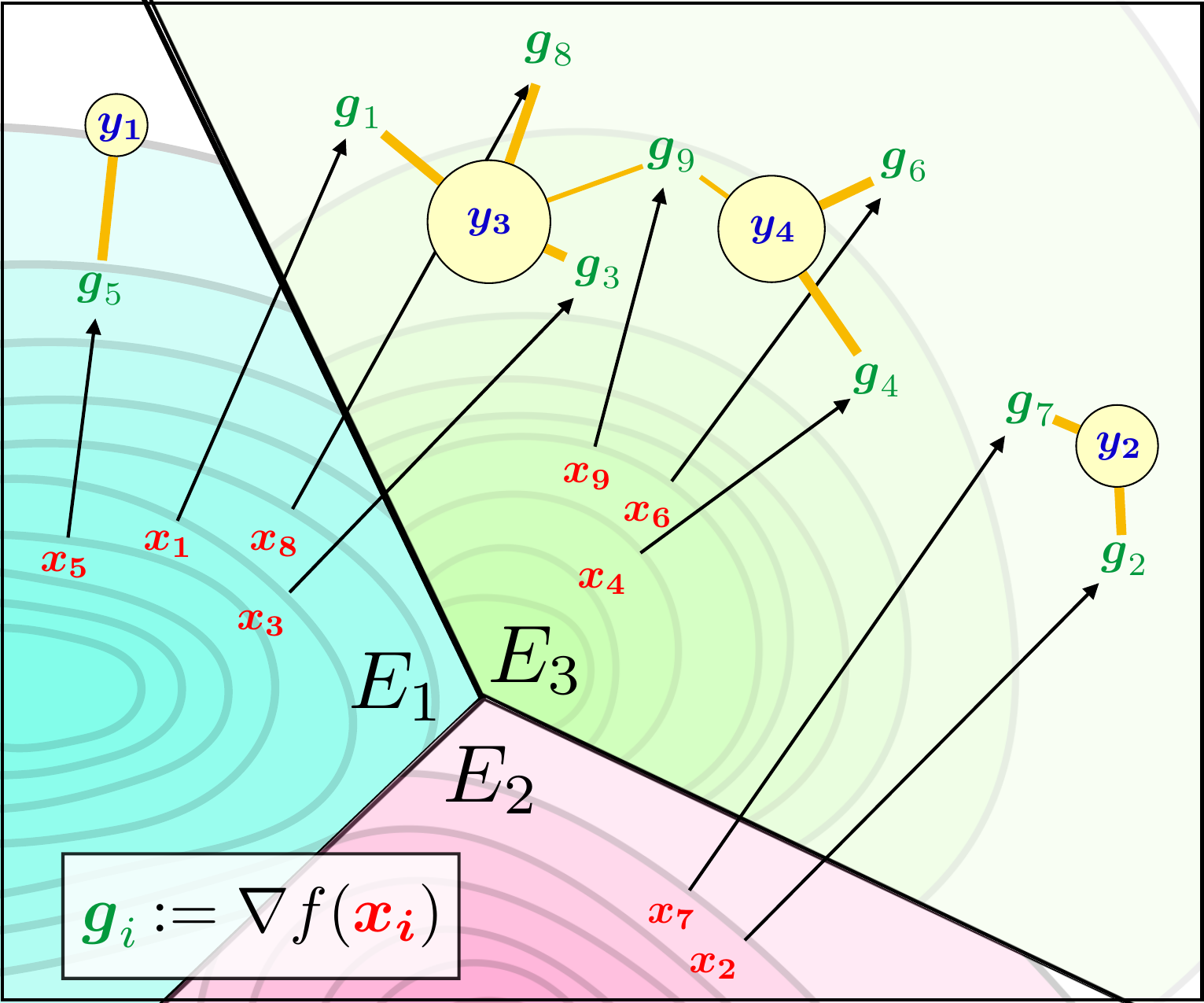}
    \caption{Points $x_i$ mapped onto points $g_i:=\nabla f(x_i)$ for a function $f$ that is locally smooth strongly convex. SSNB potentials are such that the measure of endpoints $g_i$ is as close as possible (in $\W_2$ sense) to the measure supported on the $y_j$. Here this would be the sum of the squares of the length of these orange sticks.}
    \label{fig:figdelamort}
\end{figure}


\textbf{Near-Brenier Smooth Strongly Convex Potentials.}
We will seek functions $f$ that are $\scvx$-strongly convex and $L$-smooth (or, alternatively, locally so) while at the same time such that $\nabla f_\sharp \mu$ is as \textit{close as possible} to the target $\nu$. If $\nabla f_\sharp \mu$ were to be exactly equal to $\nu$, such a function would be called a Brenier potential. We quantify that nearness in terms of the Wasserstein distance between the push-foward of $\mu$ and $\nu$ to define:

\begin{definition} \label{def:SSNB}
    Let $\mathcal{E}$ be a partition of $\Rd$ and $0 \leq \scvx \leq L$. For $\mu, \nu \in \PdRd$, we call $\opt{f}$ a ($\mathcal{E}$-locally) $L$-smooth $\scvx$-strongly convex nearest Brenier (SSNB) potential between $\mu$ and $\nu$ if
    \[
        \opt{f} \in \argmin_{f \in \mathcal{F}_{\scvx, L,\mathcal{E}}} \W_2\left(\nabla f_\sharp\mu, \nu\right).
    \]
\end{definition}


\begin{remark}
    The existence of an SSNB potential is proved in the supplementary material. When $\mathcal{E}=\{\Rd\}$, the gradient of any SSNB potential $\opt{f}$ defines an optimal transport map between $\mu$ and ${\nabla \opt{f}}_\sharp\mu$. The associated transport value $\W_2^2\left(\mu, {\nabla \opt{f}}_\sharp\mu\right)$ does not define a metric between $\mu$ and $\nu$ because it is not symmetric, and $\W_2(\mu, {\nabla \opt{f}}_\sharp\mu)=0 \not \Rightarrow \mu=\nu$ (take any $\nu$ that is not a Dirac and $\mu = \delta_{\expectation[\nu]}$).
    For more general partitions $\mathcal{E}$ one only has that property locally, and $\opt{f}$ can therefore be interpreted as a piecewise convex potential, giving rise to piecewise optimal transport maps, as illustrated in Figure~\ref{fig:figdelamort}.
\end{remark}

\textbf{Algorithmic Formulation as an Alternate QCQP/Wasserstein Problem.} We will work from now on with two discrete measures $\mu = \sum_{i=1}^n a_i \delta_{x_i}$ and $\nu = \sum_{j=1}^m b_j \delta_{y_j}$, with supports defined as $x_1, \ldots x_n \in \Rd$, $y_1, \ldots y_m \in \Rd$, and $\ba=(a_1,\dots,a_n)$ and $\bb=(b_1,\dots,b_m)$ are probability weight vectors. We write $\mathcal{U}(\ba,\bb)$ for the transportation polytope with marginals $\ba$ and $\bb$, namely the set of $n\times m$ matrices with nonnegative entries such that their row-sum and column-sum are respectively equal to $\ba$ and $\bb$. Set a desired smoothness $L > 0$ and strong-convexity parameter $\scvx \leq L$, and choose a partition $\mathcal{E}$ of $\Rd$ (in our experiments $\mathcal{E}$ is either $\{\Rd\}$, or computed using a $K$-means partition of $\supp{\mu}$). For $k \in \range{K}$, we write $I_k = \left\{ i \in \range{n} \text{ s.t. } x_i \in E_k \right\}$. The infinite dimensional optimization problem introduced in Definition~\ref{def:SSNB} can be reduced to a QCQP that only focuses on the values and gradients of $f$ at the points $x_i$. This result follows from the literature in the study of first order methods, which considers optimizing over the set of convex functions with prescribed smoothness and strong-convexity constants (see for instance~\citep[Theorem 3.8 and Theorem 3.14]{taylor2017convex}). We exploit such results to show that an SSNB $f$ can not only be estimated at those points $x_i$, but also more generally recovered at any arbitrary point in $\Rd$.

\begin{theorem}\label{thm:regasreg:optpb}
 The $n$ values $u_i:=f(x_i)$, and gradients $z_i:=\nabla f(x_i)$ of a SSNB potential $f\in\mathcal{F}_{\scvx, L,\mathcal{E}}$ can be recovered as:
 \begin{align} \label{eqn:SSNB}
     &\min_{\substack{z_1, \ldots z_n \in \Rd\\u \in \R^n}} \W_2^2 \left( \sum_{i=1}^n a_i \delta_{z_i}, \nu \right) := \min_{P\in\mathcal{U}(a,b)} \sum_{i,j} P_{ij} \|z_i - y_j\|^2\\
     &\textrm{s.t. } \forall k\leq K, \forall i,j\in I_k, \nonumber\\
     & \qquad u_i \geq u_j + \dotp{z_j}{x_i - x_j} \nonumber\\
     &\qquad \qquad + \frac{1}{2(1 - \scvx/L)} \left(
             \frac{1}{L} \|z_i - z_j\|^2 + \scvx\|x_i - x_j\|^2 \right. \nonumber \\
     &\qquad \qquad \qquad \qquad \qquad
             \left. - 2\frac{\scvx}{L} \dotp{z_j - z_i}{x_j - x_i}
         \right).\nonumber
 \end{align}
 Moreover, for $x \in E_k$, $v := f(x)$ and $g := \nabla f(x)$ can be recovered as:
 \begin{align}\label{eqn:newpoint}
     \min_{v \in \R,\, g \in \Rd} v \\
     \begin{split}
         \mathrm{\quad s.t.} \, \forall i \in I_k, v \geq & \text{ } u_i + \dotp{z_i}{x - x_i} \\
         &+ \frac{1}{2(1-\scvx/L)} \left(
             \frac{1}{L} \|g - z_i\|^2 \right.\\
         &\left. + \scvx \|x - x_i\|^2 - 2 \frac{\scvx}{L} \dotp{z_i - g}{x_i - x}
         \right).
    \end{split} \nonumber
\end{align}
\end{theorem}


We refer to the supplementary material for the proof.

We provide algorithms to compute a SSNB potential in dimension $d \geq 2$ when $\mu,\nu$ are discrete measures. In order to solve Problem~\eqref{eqn:SSNB}, we will alternate between minimizing over $(z_1, \ldots, z_n, u)$ and computing a coupling $P$ solving the OT problem. The OT computation can be efficiently carried out using Sinkhorn's algorithm~\citep{CuturiSinkhorn}. The other minimization is a convex QCQP, separable in $K$ smaller convex QCQP that can be solved efficiently. We use the barycentric projection (see Definition~\ref{def:barycentricproj} below) of $\mu$ as an initialization for the points $z$.
\section{ONE-DIMENSIONAL CASE AND THE LINK WITH CONSTRAINED ISOTONIC REGRESSION}


We consider first SSNB potentials in arguably the simplest case, namely that of distributions on the real line. We use the definition of the ``barycentric projection'' of a coupling~\cite[Def.5.4.2]{ambrosio2006gradient}, which is the most geometrically meaningful way to recover a map from a coupling.
\begin{definition}[Barycentric Projection]\label{def:barycentricproj}
    Let $\mu, \nu \in \PdRd$, and take $\pi$ an optimal transport plan between $\mu$ and $\nu$. The barycentric projection of $\pi$ is defined as the map $\barycentric{\pi} : x \mapsto \expectation_{(X,Y)\sim\pi}[Y | X=x]$.
\end{definition}

Theorem 12.4.4 in~\citep{ambrosio2006gradient} shows that $\barycentric{\pi}$ is the gradient a convex function. It is then admissible for the SSNB optimization problem defined in Theorem~\ref{thm:regasreg:optpb} as soon as it verifies regularity (Lipschitzness) and curvature (strongly increasing). Although the barycentric projection map is not optimal in general, the following proposition shows that it is however optimal for univariate measures:
\begin{proposition}\label{prop:oneD:isotonic}
    Let $\mu, \nu \in \PdR$ and $0 \leq \scvx \leq L$. Suppose $\mu \continuouswrt \Lebesgue{1}$, or is purely atomic. Then the set of SSNB potentials between $\mu$ and $\nu$ is the set of solutions to
    \[
        \min_{f \in \mathcal{F}_{\scvx,L,\mathcal{E}}} \| f' - \barycentric{\pi} \|^2_{L^2(\mu)}
    \]
    where $\pi$ is the unique optimal transport plan between $\mu$ and $\nu$ given by~\citep[Theorem 2.9]{SantambrogioBook}.
\end{proposition}

\begin{figure}[!h]
    \includegraphics[width=0.48\textwidth]{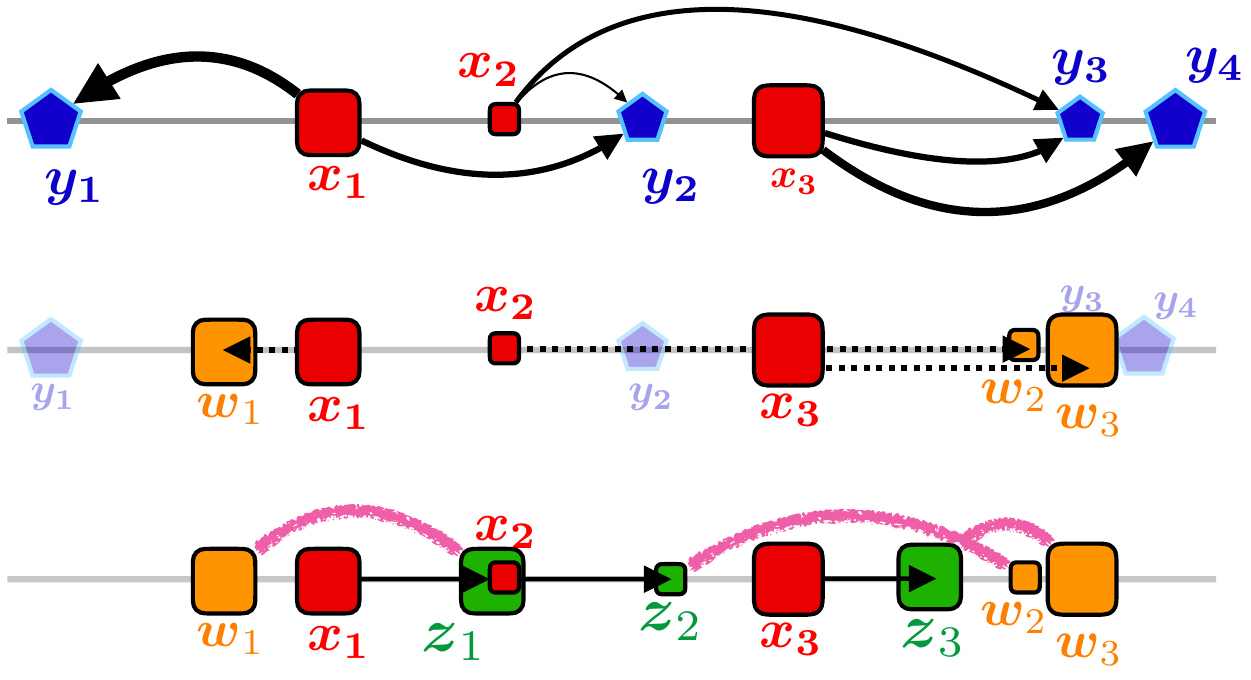}
    \caption{Top: optimal transport between two discrete measures $\mu,\nu$. Middle: the barycentric projection $\bw$ of points $\bx$ is displayed and corresponds to a Monge map (no mass splitting). Considering here for instance $\scvx=0.5$ and $L=1$, the map that associates $x_i$ to $w_i$ is not $1$-Lipschitz at pairs $(1,2)$ or $(1,3)$ and over-contracting in pair $(2,3)$. Bottom: To compute points $z_i$ that minimize their transport cost to the $w_i$ (pink curves) while still ensuring $x_i\mapsto z_i$ is $L$-Lipschitz and strongly increasing amounts to solving the $L$-Lipschitz $\scvx$-strongly increasing isotonic regression problem~\eqref{eq:Lisotonic}.}
    \label{fig:Lisotonic}
\end{figure}

\textbf{Discrete Computations.} Suppose $\mu = \sum_{i=1}^n a_i \delta_{x_i}$ is discrete with $x_1 \leq \ldots \leq x_n$, and $\nu$ is arbitrary. Let us denote by $Q_\nu$ the (generalized) quantile function of $\nu$. Writing $\pi$ for the optimal transport plan between $\mu$ and $\nu$, the barycentric projection $\barycentric{\pi}$ is explicit. Writing  $\alpha_0 := 0$, $\alpha_i := \sum_{k=1}^i a_k$, one has $\barycentric{\pi}(x_i) = \frac{1}{a_i} \int_{\alpha_{i-1}}^{\alpha_i} Q_\nu(t) \,dt$  (proof in the supplementary material).


If $\nu$ is also discrete, with weights $\mathbf{b}=(b_1,\dots,b_m)$ and sorted support $\mathbf{y}=(y_1,\dots,y_m)\in\R^m$, where $y_1\leq\dots\leq y_m$, one can recover the coordinates of the vector $(\barycentric{\pi}(x_i))_i$ of barycentric projections as  
$$\bw:= \text{diag}(\mathbf{a}^{-1})\mathbf{NW}(\mathbf{a},\mathbf{b})\,\mathbf{y},$$ 
where $\mathbf{NW}(\mathbf{a},\mathbf{b})$ is the so-called \textit{North-west corner} solution~\cite[\S3.4.2]{MAL-073} obtained in linear time w.r.t $n,m$ by simply filling up greedily the transportation matrix from top-left to down-right.  We deduce from Proposition~\ref{prop:oneD:isotonic} that a SSNB potential can be recovered by solving a weighted (and local, according to the partition $\mathcal{E}$) constrained isotonic regression problem (see Fig.~\ref{fig:Lisotonic}):
\vskip-0.8cm
\begin{align}\label{eq:Lisotonic}
    &\min_{z \in \R^n} \sum_{i=1}^n a_i (z_i - w_i)^2\\
    &\text{s.t. } \forall k\leq K,\, \forall i,i+1 \in I_k, \nonumber\\
    & \qquad \quad \scvx(x_{i+1}-x_i) \leq z_{i+1}-z_i \leq L(x_{i+1}-x_i). \nonumber
\end{align}
The gradient of a SSNB potential $\opt{f}$ can then be retrieved by taking an interpolation of $x_i \mapsto z_i$ that is piecewise affine.


Algorithms solving the Lipschitz isotonic regression were first designed by~\cite{yeganova2009isotonic} with a $\mathcal{O}(n^2)$ complexity. \citep{agarwal2010lipschitz, kakade2011efficient} developed $\mathcal{O}(n \log{n})$ algorithms. A Smooth NB potential can therefore be exactly computed in $\mathcal{O}(n \log{n})$ time, which is the same complexity as of optimal transport in one dimension. Adding up the strongly increasing property, Problem~\eqref{eq:Lisotonic} can also be seen as least-squares regression problem with box constraints. Indeed, introducing $m$ variables $v_i\geq 0$, and defining $z_i$ as the partial sum $\bv$, namely $z_i=\sum_{j=1}^i v_j$ (or equivalently $v_i=z_{i}-z_{i-1}$ with $z_{0}:=0$), and writing $u^-_i=\scvx(x_{i+1}-x_{i})$, $u^+_i=L(x_{i+1}-x_{i})$ one aims to find $\bv$ that minimizes $\|A\bv-\bw\|_a^2$ s.t. $\buminus\leq \bv\leq \bu$, where $A$ is the lower-triangular matrix of ones and $\|\cdot\|_a$ is the Euclidean norm weighted by $a$. In our experiments, we have found that a projected gradient descent approach to solve this problem performed in practice as quickly as more specialized algorithms and was easier to parallelize when comparing a measure $\mu$ to several measures $\nu$.

\begin{figure}[!h]
    \includegraphics[width=0.48\textwidth]{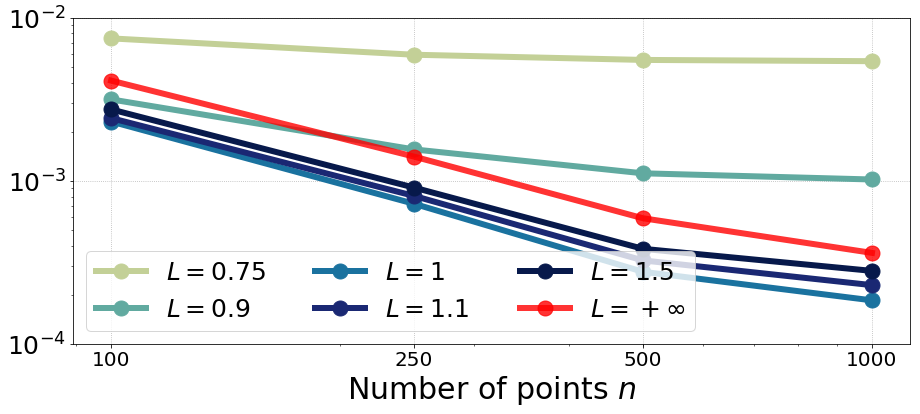}
    \caption{Take measures $\mu = \nu$ be the uniform measure over $[0,1]$. For several $n$, we consider $\hat\mu_n,\hat\nu_n$ empirical measures over $n$ iid samples from $\mu,\nu$, from which we compute a SSNB potential $\hat f_n$ with different values $L$, and $\scvx = \min\{1, L\}$ (and $\scvx=0$ if $L=\infty$). We plot the estimation error $|\W_2^2(\hat\mu_n, \hat\zeta_n) - \W_2^2(\mu, \nu)|$ depending on $n$ and $L$, averaged over $100$ runs, where $\hat\zeta_n = \hat {f_n}_\sharp \mu_n$. If $L \leq \lip{\mathrm{Id}} = 1$, the error does not converge to $0$. Otherwise, the convergence is faster when $L$ is closer to $1$. The case $L=\infty$ corresponds to the classical OT estimator $\hat\zeta_n = \hat\nu_n$.}
    \label{fig:onedimension_values}
\end{figure}
\section{ESTIMATION OF WASSERSTEIN DISTANCE AND MONGE MAP}


Let $\mu, \nu \in \PdRd$ be two compactly supported measures with densities w.r.t the Lebesgue measure in $\Rd$. Brenier theorem gives the existence of an optimal Brenier potential $\opt{f}$, \textit{i.e.} $\opt{f}$ is convex and $\nabla \opt{f}_\sharp\mu = \nu$. Our goal is twofold: estimate the map $\nabla \opt{f}$ and the value of $\W_2(\mu, \nu)$ from samples.

Let $x_1, \ldots, x_n \sim \mu$ and $y_1, \ldots, y_n \sim \nu$ be i.i.d samples from the measures, and define $\hat\mu_n := \frac{1}{n}\sum_{i=1}^n \delta_{x_i}$ and $\hat\nu_n := \frac{1}{n}\sum_{i=1}^n \delta_{y_i}$ the empirical measures over the samples.

Let $\hat f_n$ be a SSNB potential between $\hat\mu_n$ and $\hat\nu_n$ with $\mathcal{E}=\{\Rd\}$. Then for $x \in \supp{\mu}$, a natural estimator of $\nabla \opt{f}(x)$ is given by a solution $\nabla \hat f_n(x)$ of~\eqref{eqn:newpoint}. This defines an estimator $\nabla \hat f_n: \Rd \to \Rd$ of $\nabla \opt{f}$, that we use to define a plug-in estimator for $\W_2(\mu,\nu)$:
\begin{definition}
    We define the SSNB estimator of $\W_2^2(\mu,\nu)$ as $\widehat \W_2^2(\mu,\nu) := \int \|x - \nabla \hat f_n(x)\|^2 \,d\mu(x)$.
\end{definition}\vspace{-0.3cm}
Since $\nabla \hat f_n$ is the gradient of a convex Brenier potential when $\mathcal{E}=\{\R^d\}$, it is optimal between $\mu$ and ${\nabla \hat f_n}_\sharp\mu$ and $\widehat \W_2(\mu,\nu) = \W_2(\mu, {\nabla \hat f_n}_\sharp\mu)$. If $\mathcal{E} \neq \{\Rd\}$, $\nabla \hat f_n$ is the gradient of a locally convex Brenier potential, and is not necessarily globally optimal between $\mu$ and ${\nabla \hat f_n}_\sharp\mu$. In that case $\widehat \W_2(\mu,\nu)$ is an approximate upper bound of $\W_2(\mu, {\nabla \hat f_n}_\sharp\mu)$. In any case, the SSNB estimator can be computed using Monte-Carlo integration, whose estimation error does not depend on the dimension $d$.

\begin{algorithm}[!h]
	\caption{Monte-Carlo approximation of the SSNB estimator}
	\begin{algorithmic}
   		\STATE {\bfseries Input:} $\hat\mu_n$, $\hat\nu_n$, partition $\mathcal{E}$, number $N$ of Monte-Carlo samples
		\STATE . $(u_i, z_i)_{i\leq n} \leftarrow$ solve SSNB~\eqref{eqn:SSNB} between $\hat\mu_n$, $\hat\nu_n$
   		\FOR{$j \in \range{N}$}
		\STATE . Draw $\hat x_j \sim \mu$
        \STATE . Find $k$ s.t. $\hat x_j \in E_k$ (k-means)
        \STATE . $\nabla \hat f_n(\hat x_j) \leftarrow$ solve QCQP~\eqref{eqn:newpoint}
  		\ENDFOR
		\STATE{\bfseries Output:} $\widehat W = \left[(1/N)\sum_{j=1}^N \|\hat x_j - \nabla \hat f_n(\hat x_j)\|^2 \right]^{1/2}$
	\end{algorithmic}
\end{algorithm}

We show that when the Brenier potential $\opt{f}$ is globally regular, the SSNB estimator is strongly consistent:
\begin{proposition}\label{prop:estimation:L2}
Choose $\mathcal{E} = \{\Rd\}$, $0 \leq \scvx \leq L$. If $\opt{f} \in \mathcal{F}_{\scvx, L, \mathcal{E}}$, it almost surely holds:
\[
    \left\vert \W_2(\mu, \nu) - \widehat \W_2(\mu,\nu) \right\vert
    \underset{n \to \infty}{\longrightarrow} 0.
\]
\end{proposition}
The study of the theoretical rate of convergence of this estimator is beyond the scope of this work. Recent results~\citep{hutter2019minimax,flamary2019concentration} show that assuming some regularity of the Monge map $\nabla \opt{f}$ leads to improved sample complexity rates. Numerical simulations (see section~\ref{sec:exp_stat}) seem to indicate a reduced estimation error for SSNB over the classical $\W_2(\hat\mu_n, \hat\nu_n)$, even in the case where $\nabla \opt{f}$ is only locally Lipschitz and $\mathcal{E} \neq \{\Rd\}$. If $L < \lip{\nabla \opt{f}}$, the SSNB estimator $\widehat\W_2(\mu,\nu)$ is not consistent, as can be seen in Figure~\ref{fig:onedimension_values}.
\section{EXPERIMENTS}
All the computations were performed on a i9 2,9 GHz CPU, using MOSEK as a convex QCQP solver.

\subsection{Numerical Estimation of Wasserstein Distances and Monge Maps}\label{sec:exp_stat}

\begin{figure}[!h]
    \centering
    \includegraphics[width=0.20\textwidth]{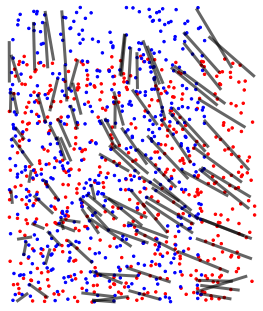} \\
    \includegraphics[width=0.40\textwidth]{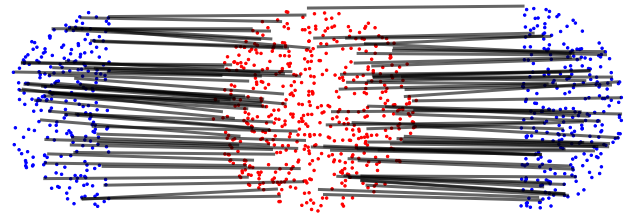}
    \caption{In the global (top) and local (bottom) regularity settings with $d=2$, we plot $n=500$ i.i.d samples from $\mu$ (red) and $\nu$ (blue). Black lines show the displacements $x \mapsto \nabla \hat f_n(x)$ for some new points $x \in \supp(\mu) \setminus \supp(\hat\mu_n)$, computed by solving QCQP~\eqref{eqn:newpoint}.
    }\label{fig:figures_semiball_cube}
\end{figure}

In this experiment, we consider two different settings:\newline
\textbf{Global regularity:} $\mu$ is the uniform measure over the unit cube in $\Rd$, and $\nu = T_\sharp\mu$ where $T(x) = \Omega_d x$, where $\Omega_d$ is the diagonal matrix with diagonal coefficients: $0.8 + \frac{0.4}{d-1}(i-1)$, for $i \in \range{d}$. $T$ is the gradient of the convex function $f: x \mapsto \frac{1}{2}\|\Omega^{1/2}x\|^2$, so it is the optimal transport map. $f$ is globally $\scvx=0.8$-strongly convex and $L=1.2$-smooth.\newline
\textbf{Local Regularity:} $\mu$ is the uniform measure over the unit ball in $\Rd$, and $\nu = T_\sharp\mu$ where $T(x_1,\ldots,x_d) = (x_1 + 2\sign(x_1), x_2, \ldots, x_d)$. As can be seen in Figure~\ref{fig:figures_semiball_cube} (bottom), $T$ splits the unit ball into two semi-balls. $T$ is a subgradient of the convex function $f: x \mapsto \frac{1}{2}\|x\|^2 + 2|x_1|$, so it is the optimal transport map. $f$ is $\scvx=1$-strongly convex, but is not smooth: $\nabla f$ is not even continuous. However, $f$ is $L=1$-smooth by parts.

For each of those two settings, we consider i.i.d samples $x_1, \ldots, x_n \sim \mu$ and $y_1,  \ldots, y_n \sim \nu$ for different values of $n \in \N$, and denote by $\hat\mu_n$ and $\hat\nu_n$ the respective empirical measures on these points. Given a number of clusters $1 \leq K \leq n$, we compute the partition $\mathcal{E} = \{E_1, \ldots, E_K\}$ by running k-means with $K$ clusters on data $x_1, \ldots, x_n$. In both settings, we run the algorithms with $\hat\scvx=0.6$ and $\hat L=1.4$. We give experimental results on the statistical performance of our SSNB estimator, computed using Monte-Carlo integration, compared to the classical optimal transport estimator $\W_2(\hat\mu_n,\hat\nu_n)$. This performance depends on three parameters: the number $n$ of points, the dimension $d$ and the number of clusters $K$.

\begin{figure}[h!]
    \includegraphics[width=0.48\textwidth]{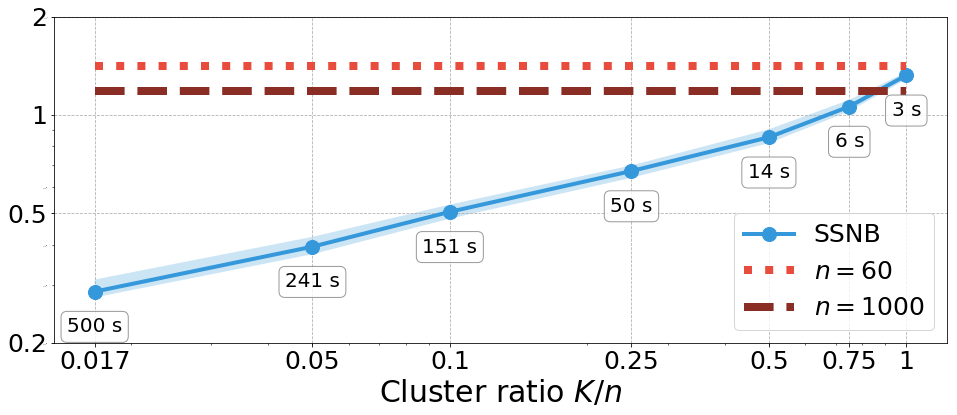}
    \includegraphics[width=0.48\textwidth]{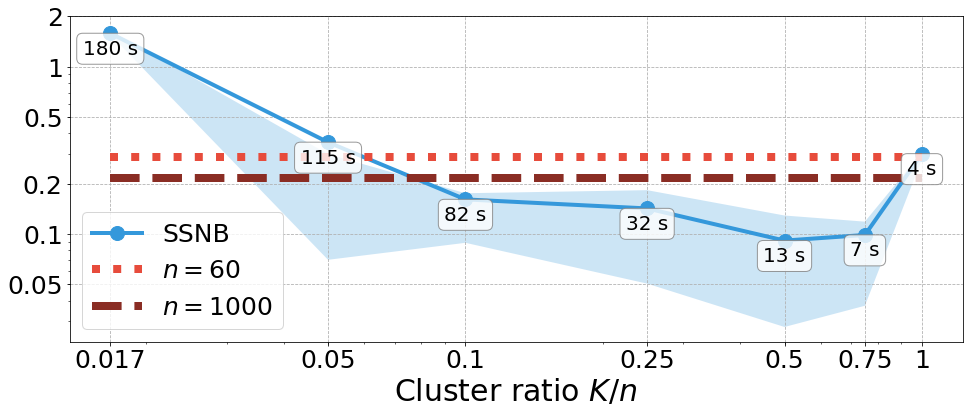}
    \caption{In the global (top) and local (bottom) regularity settings, we plot (on a log-log scale) the estimation error $\vert \W_2(\mu,\nu) - \widehat\W_2(\mu,\nu) \vert$ (blue line) depending on the cluster ratio $K/n$, with fixed number of points $n=60$, dimension $d=30$ and Monte-Carlo samples $N=50$. The curves/shaded areas show the mean error/25\%-75\% percentiles over $20$ data samples. The bubbles show the mean running time for the SSNB estimator computation. We plot the classical estimation error for $n=60$ (light red dotted) and $n=1000$ (dark red dashed) for comparison.}\label{fig:dependence_on_K}
\end{figure}

In Figure~\ref{fig:dependence_on_K}, we plot the estimation error depending on the cluster ratio $K/n$ for fixed $n=60$ and $d=30$. In the global regularity setting (Figure~\ref{fig:dependence_on_K} top), the error seems to be exponentially increasing in $K$, whereas the computation time decreases with $K$: there is a trade-off between accuracy and computation time. In the local regularity setting (Figure~\ref{fig:dependence_on_K} bottom), the estimation error is large when the number of clusters is too small (because we ask for too much regularity) or too large. Interestingly, the number of clusters can be taken large and still leads to good results. In both settings, even a large number of clusters is enough to outperform the classical OT estimator, even when SSNB is computed with a much smaller number of points. Note that when $K/n = 1$, the SSNB estimator is basically equivalent (up to the Monte-Carlo integration error) to the classical OT estimator.

\begin{figure}[h!]
    \includegraphics[width=0.48\textwidth]{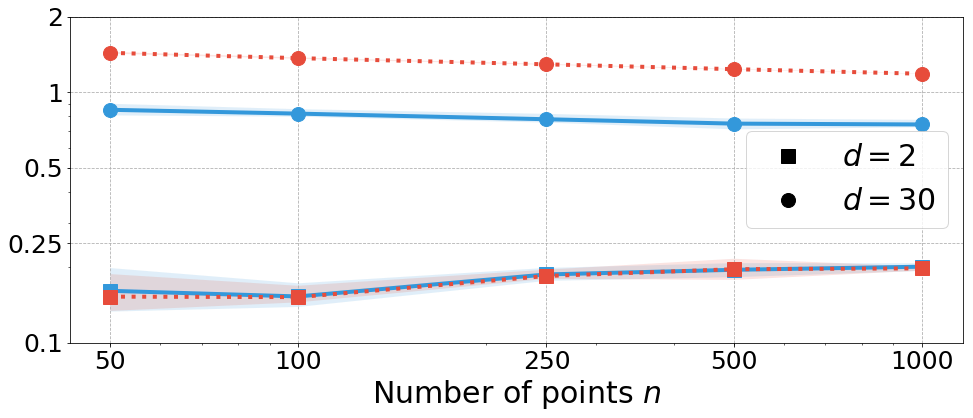}
    \includegraphics[width=0.48\textwidth]{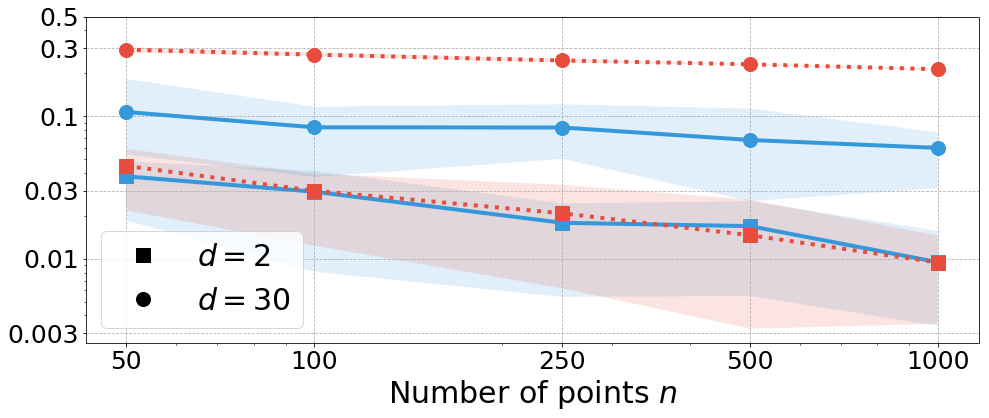}
    \caption{In the global (top) and local (bottom) regularity settings, we plot the estimation error of SSNB $\vert \W_2(\mu,\nu) - \widehat\W_2(\mu,\nu) \vert$ (blue line) and classical OT estimator $\vert \W_2(\mu,\nu) - \W_2(\hat\mu_n,\hat\nu_n) \vert$ (red dotted) depending on the number of points $n \in \{50, 100, 250, 500, 1000\}$, for dimension $d \in \{2, 30\}$. Here, the cluster ratios are taken constant equal to $0.5$ (resp. $0.75$) for the global (resp. local) regularity experiment. The number of Monte-Carlo samples is $N=50$. The curves/shaded areas show the mean error/25\%-75\% percentiles over $20$ data samples.}\label{fig:dependence_on_n}
\end{figure}

In Figure~\ref{fig:dependence_on_n}, we plot the estimation error depending on the number of points $n$, for fixed cluster ratio $K/n$ and different dimension $d \in \{2, 30\}$. In both settings, and for both low ($d=2$) and high ($d=30$) dimension, the SSNB estimator seem to have the same rate as the classical OT estimator, but a much better constant in high dimension. This means that in high dimension, the SSNB estimator computed with a small number of points can be much more accurate that the classical OT estimator computed with a large number of points.

\subsection{Domain Adaptation}

Domain adaptation is a way to transfer knowledge from a source to a target dataset. The source dataset consists of labelled data, and the goal is to classify the target data. \citet{courty2016optimal, perrot2016mappingestimation} proposed to use optimal transport (and different regularized version of OT) to perform such a task: the OT map from the source dataset to the target dataset (seen as empirical measures over the source/target data) is computed. Then each target data is labelled  according to the label of its nearest neighbor among the transported source data.

The Caltech-Office datasets A, C, D, W contain images of ten different objects coming from four different sources: Amazon, Caltech-256, DSLR and Webcam. We consider DeCAF6 features~\citep{donahue2014decaf}, which are sparse $4096$-dimensional vectors. For each pair of source/target datasets, both datasets are cut in half (train and test): hyperparameters are learnt on the train half and we report the mean ($\pm$ std) test accuracy over 10 random cuts, computed using a $1$-Nearest Neighbour classifier on the transported source data. Results for OT, entropic OT, Group-Lasso and entropy regularized OT and SSNB are given in Table~\ref{table:DA}.

In order to compute the SSNB mapping, we \textit{a)} quantize the source using k-means in each class with $k=4$ centroids, \textit{b)} learn the SSNB map between the $40$ centroids and the target by solving~\eqref{eqn:SSNB}, \textit{c)} transport the source data by solving~\eqref{eqn:newpoint}.

\begin{table}[!h]
    \centering
    \resizebox{0.48\textwidth}{!}{
        \begin{tabular}{|c|c|c|c|c|}
            \hline
            \textbf{Domains} & \textbf{OT} & \textbf{OT-IT} & \textbf{OT-GL} & \textbf{SSNB} \\ \hline
            \textbf{A} $\boldsymbol{\rightarrow}$ \textbf{C} & $74.0$ & $81.0\pm2.2$ & $\boldsymbol{86.6\pm1.0}$ & $83.9\pm1.6$ \\ \hline
            \textbf{A} $\boldsymbol{\rightarrow}$ \textbf{D} & $65.0$ & $\boldsymbol{84.3\pm5.2}$ & $\boldsymbol{82.9\pm5.2}$ & $\boldsymbol{79.2\pm3.7}$ \\ \hline
            \textbf{A} $\boldsymbol{\rightarrow}$ \textbf{W} & $65.0$ & $\boldsymbol{79.3\pm5.3}$ & $\boldsymbol{81.6\pm3.1}$ & $\boldsymbol{82.3\pm3.1}$ \\ \hline
            \textbf{C} $\boldsymbol{\rightarrow}$ \textbf{A} & $75.3$ & $90.3\pm1.3$ & $\boldsymbol{91.1\pm1.8}$ & $\boldsymbol{91.9\pm1.1}$ \\ \hline
            \textbf{C} $\boldsymbol{\rightarrow}$ \textbf{D} & $65.6$ & $\boldsymbol{83.9\pm4.0}$ & $\boldsymbol{85.8\pm2.3}$ & $81.1\pm4.9$ \\ \hline
            \textbf{C} $\boldsymbol{\rightarrow}$ \textbf{W} & $64.8$ & $\boldsymbol{77.1\pm3.9}$ & $\boldsymbol{81.9\pm5.1}$ & $\boldsymbol{79.3\pm2.8}$ \\ \hline
            \textbf{D} $\boldsymbol{\rightarrow}$ \textbf{A} & $69.6$ & $88.3\pm3.1$ & $\boldsymbol{90.5\pm1.9}$ & $\boldsymbol{91.2\pm0.8}$ \\ \hline
            \textbf{D} $\boldsymbol{\rightarrow}$ \textbf{C} & $66.4$ & $78.0\pm2.9$ & $\boldsymbol{85.6\pm2.2}$ & $81.4\pm1.9$ \\ \hline
            \textbf{D} $\boldsymbol{\rightarrow}$ \textbf{W} & $84.7$ & $\boldsymbol{96.1\pm2.1}$ & $93.8\pm2.1$ & $\boldsymbol{95.6\pm1.5}$ \\ \hline
            \textbf{W} $\boldsymbol{\rightarrow}$ \textbf{A} & $66.4$ & $82.5\pm4.3$ & $\boldsymbol{89.3\pm2.7}$ & $\boldsymbol{89.8\pm2.9}$ \\ \hline
            \textbf{W} $\boldsymbol{\rightarrow}$ \textbf{C} & $62.5$ & $74.8\pm2.3$ & $80.2\pm2.4$ & $\boldsymbol{82.7\pm1.4}$ \\ \hline
            \textbf{W} $\boldsymbol{\rightarrow}$ \textbf{D} & $87.3$ & $\boldsymbol{97.5\pm2.1}$ & $\boldsymbol{96.4\pm3.7}$ & $\boldsymbol{95.9\pm3.2}$ \\ \hdashline
            \textbf{Mean} & $70.6\pm7.8$ & $\boldsymbol{84.4\pm7.0}$ & $\boldsymbol{87.1\pm4.9}$ & $\boldsymbol{86.2\pm6.0}$ \\ \hline
        \end{tabular}
    }
    \caption{OT-IT: Entropy regularized OT. OT-GL: Entropy + Group Lasso regularized OT. Search intervals for OT-IT and OT-GL are $\epsilon, \eta \in \{10^{-3}, \ldots,10^{3}\}$ with normalized cost, and for SSNB: $\ell \in \{0.2, 0.5, 0.7, 0.9\}$, $L \in \{0.3, 0.5, 0.7, 0.9, 1.3\}$. The best results are in bold.}\label{table:DA}
\end{table}

\subsection{Color Transfer}

Given a source and a target image, the goal of color transfer is to transform the colors of the source image so that it looks similar to the target image color palette. Optimal transport has been used to carry out such a task, see \emph{e.g.}~\citep{bonneel2015sliced, ferradans2014regularized, rabin2014adaptive}. Each image is represented by a point cloud in the RGB color space identified with $[0,1]^3$. The optimal transport plan $\pi$ between the two point clouds give, up to a barycentric projection, a transfer color mapping.

It is natural to ask that similar colors are transferred to similar colors, and that different colors are transferred to different colors. These two demands translate into the smoothness and strong convexity of the Brenier potential from which derives the color transfer mapping. We therefore propose to compute a SSNB potential and map between the source and target distributions in the color space.

In order to make the computations tractable, we compute a k-means clustering with $30$ clusters for each point cloud, and compute the SSNB potential using the two empirical measures on the centroids.

We then recompute a k-means clustering of the source point cloud with $1000$ clusters. For each of the $1000$ centroids, we compute its new color by solving QCQP~\eqref{eqn:newpoint}. A pixel in the original image then sees its color changed according to the transformation of its nearest neighbor among the $1000$ centroids.

\begin{figure}[!h]
	\centering
    \captionsetup[subfigure]{justification=centering}
   	\begin{subfigure}[b]{0.15\textwidth}
	    \centering
        \includegraphics[width=\textwidth]{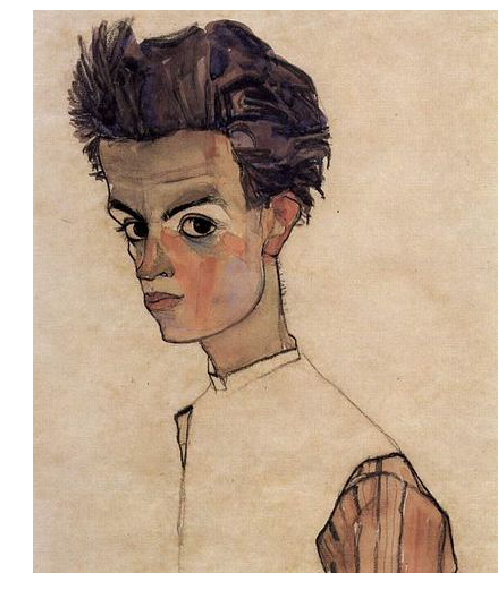}
		\caption{Original $\quad$ Image}\label{fig:schiele}
   	 \end{subfigure}
	 \begin{subfigure}[b]{0.15\textwidth}
	    \centering
		\includegraphics[width=\textwidth]{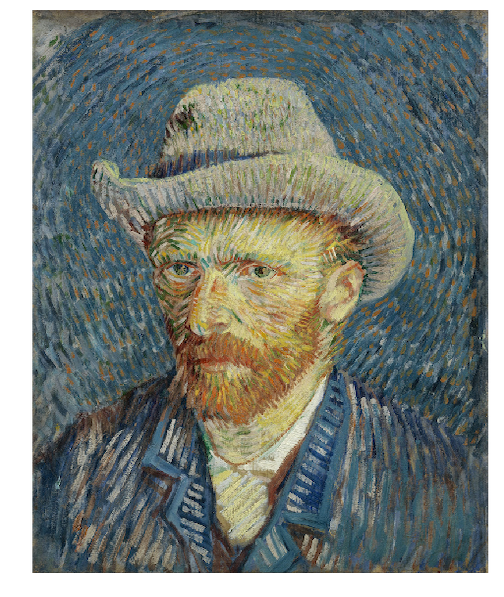}
		\caption{Target Image \phantom{Classical OT, $\W \approx 0$}}\label{fig:vangogh}
    \end{subfigure}
	\begin{subfigure}[b]{0.15\textwidth}
	    \centering
		\includegraphics[width=\textwidth]{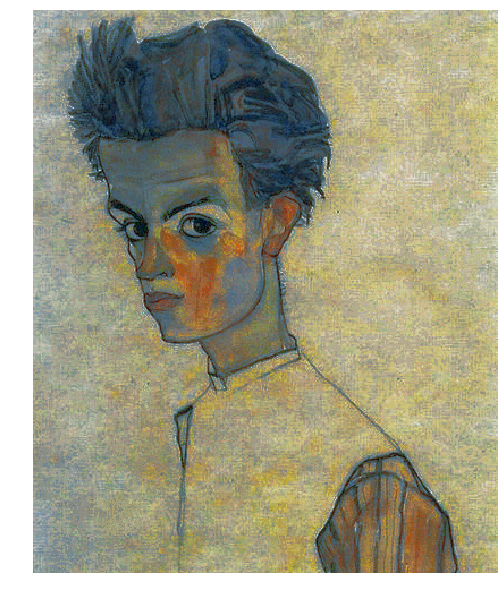}
		\caption{Classical OT, $\W \approx 0$}
    \end{subfigure}
	\begin{subfigure}[b]{0.15\textwidth}
	    \centering
		\includegraphics[width=\textwidth]{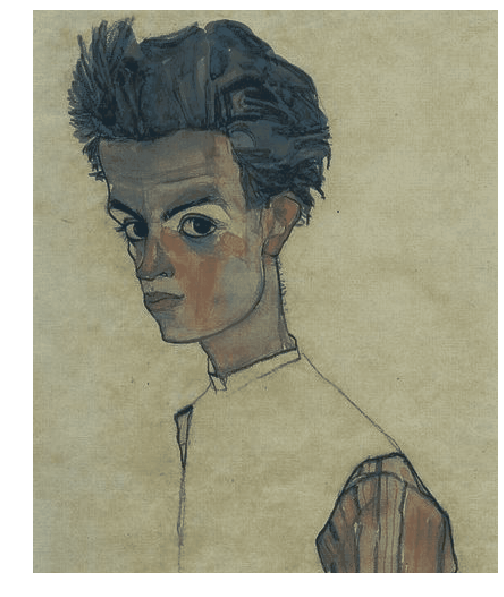}
		\caption{$\scvx = 0, L = 1$, $\W \approx 1\times 10^{-2}$}
    \end{subfigure}
	\begin{subfigure}[b]{0.15\textwidth}
	    \centering
		\includegraphics[width=\textwidth]{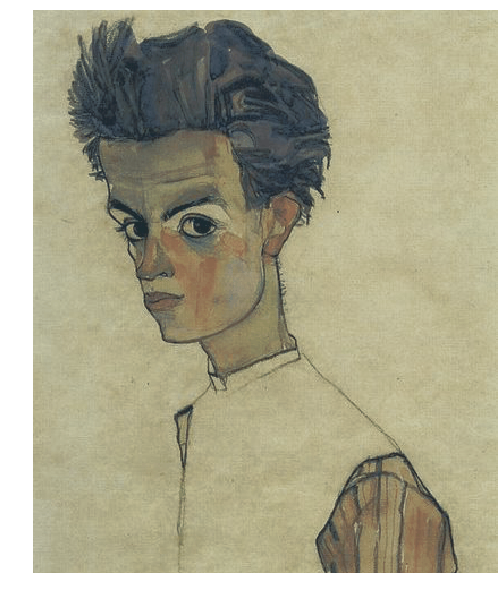}
		\caption{$\scvx = .5, L = 1$, $\W \approx 1\times 10^{-2}$}
    \end{subfigure}
	\begin{subfigure}[b]{0.15\textwidth}
	    \centering
		\includegraphics[width=\textwidth]{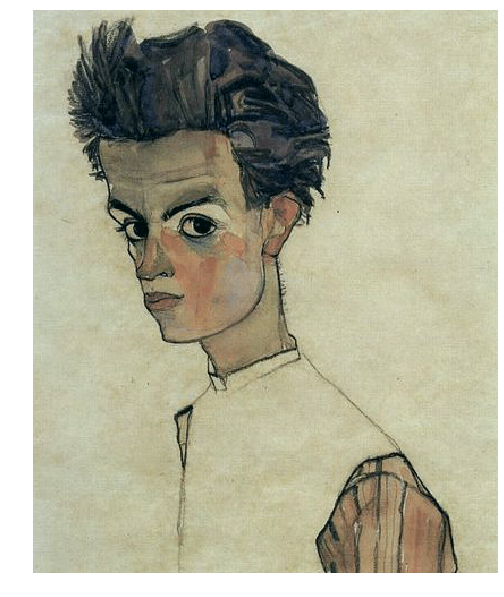}
		\caption{$\scvx = 1, L = 1$, $\W \approx 2\times 10^{-2}$}
    \end{subfigure}
	\begin{subfigure}[b]{0.15\textwidth}
	    \centering
		\includegraphics[width=\textwidth]{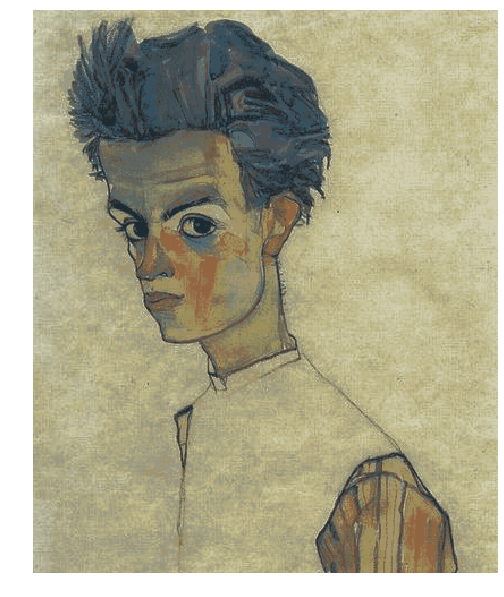}
		\caption{$\scvx = 0, L = 2$, $\W \approx 4\times 10^{-3}$}
    \end{subfigure}
	\begin{subfigure}[b]{0.15\textwidth}
	    \centering
		\includegraphics[width=\textwidth]{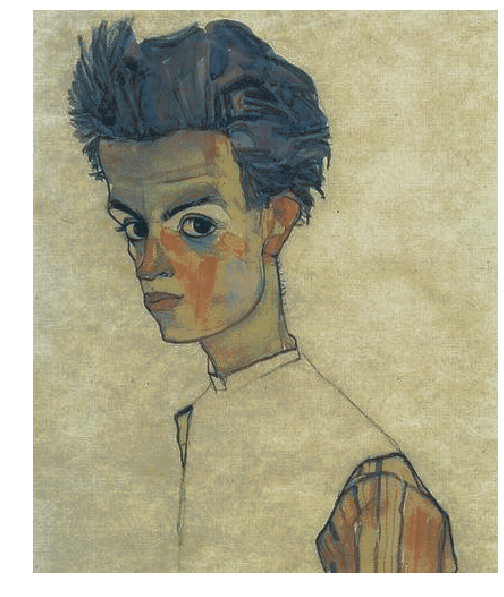}
		\caption{$\scvx = .5, L = 2$, $\W \approx 5\times 10^{-3}$}
    \end{subfigure}
	\begin{subfigure}[b]{0.15\textwidth}
	    \centering
		\includegraphics[width=\textwidth]{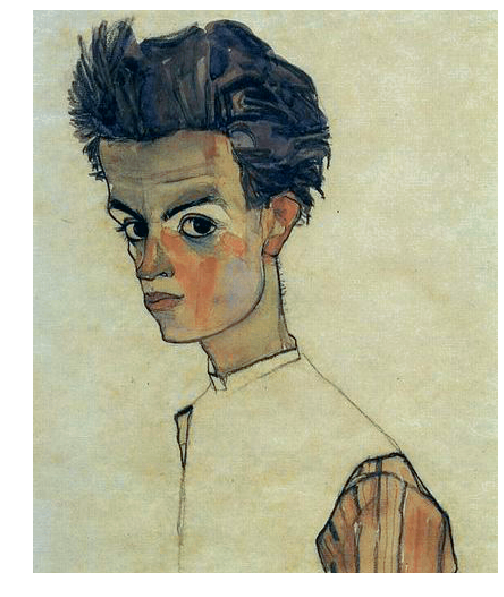}
		\caption{$\scvx = 1, L = 2$, $\W \approx 2\times 10^{-2}$}
    \end{subfigure}
	\begin{subfigure}[b]{0.15\textwidth}
	    \centering
		\includegraphics[width=\textwidth]{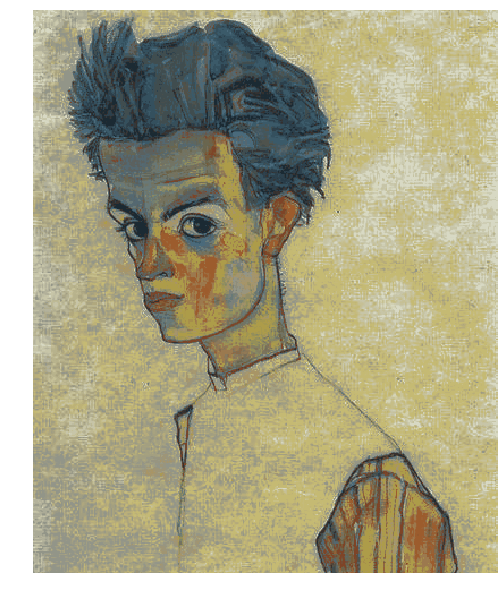}
		\caption{$\scvx = 0, L = 5$, $\W \approx 2\times 10^{-4}$}
    \end{subfigure}
	\begin{subfigure}[b]{0.15\textwidth}
	    \centering
		\includegraphics[width=\textwidth]{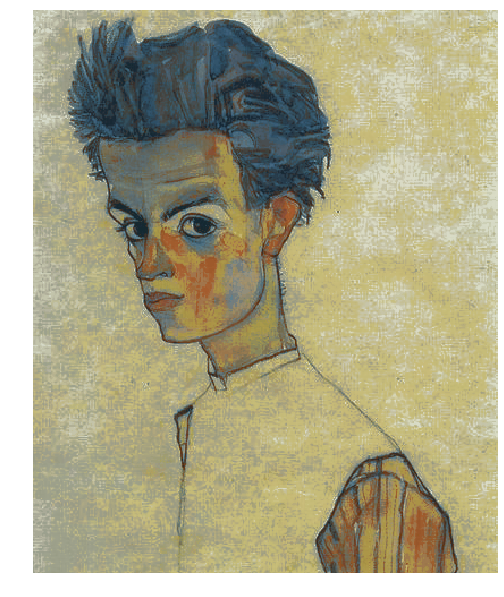}
		\caption{$\scvx = .5, L = 5$, $\W \approx 1\times 10^{-3}$}
    \end{subfigure}
	\begin{subfigure}[b]{0.15\textwidth}
	    \centering
		\includegraphics[width=\textwidth]{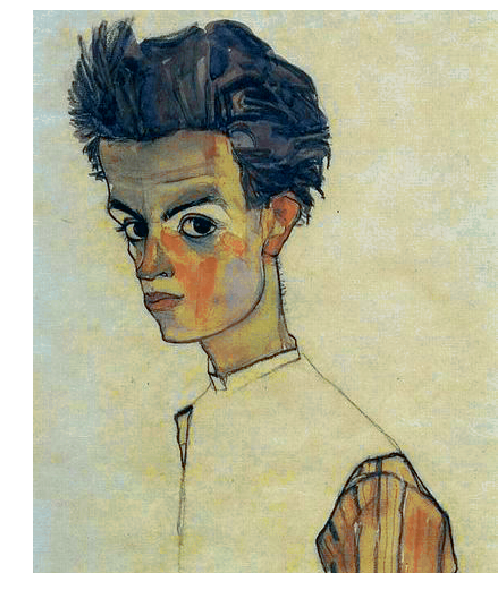}
		\caption{$\scvx = 1, L = 5$, $\W \approx 2\times 10^{-2}$}
    \end{subfigure}
   	\caption{(a) Schiele's portrait. (b) Van Gogh's portrait. (c) Color transfer using classical OT. (d)-(l) Color transfer using SSNB map, for $\scvx \in \{0, 0.5, 1\}$ and $L \in \{1, 2, 5\}$. The value $\W$ corresponds to the Wasserstein distance between the color distribution of the image and the color distribution of Van Gogh's portrait. The smaller $\W$, the greater the fidelity to Van Gogh's portrait colors.}
	 \label{fig:colortransfer}
\end{figure}

In Figure~\ref{fig:colortransfer}, we show the color-transferred results using OT, or SSNB potentials for different values of parameters $\scvx$ and $L$. Smaller values of $L$ give more uniform colors, while larger values of $\scvx$ give more contrast.

\section{CONCLUSION}
We have proposed in this work the first computational procedure to estimate optimal transport that incorporates smoothness and strongly convex (local) constraints on the Brenier potential, or, equivalently, that ensures that the optimal transport map has (local) distortion that is both upper and lower bounded. These assumptions are natural for several problems, both high and low dimensional, can be implemented practically and advance the current knowledge on handling the curse of dimensionality in optimal transport.
\vskip 0.4cm
\textbf{Acknowledgments.} We thank J. Altschuler for his remark that the QCQP~\eqref{eqn:SSNB} considered in Theorem~\ref{thm:regasreg:optpb} is in fact convex (unlike in the optimization literature where it is not) and can therefore be exactly solved and not approximated through an SDP. \newline
F.P.P and M.C. acknowledge the support of a "Chaire d'excellence de l'IDEX Paris Saclay".
A.A. is at the d\'epartement d'informatique de l'ENS, \'Ecole normale sup\'erieure, UMR CNRS 8548, PSL Research University, 75005 Paris, France, and INRIA Sierra project-team. AA would like to acknowledge support from the {\em ML and Optimisation} joint research initiative with the {\em fonds AXA pour la recherche} and Kamet Ventures, a Google focused award, as well as funding by the French government under management of Agence Nationale de la Recherche as part of the "Investissements d'avenir" program, reference ANR-19-P3IA-0001 (PRAIRIE 3IA Institute).

\bibliographystyle{icml2020}
\bibliography{biblio.bib}

\clearpage
\appendix
\section{PROOFS}
\paragraph{Proof for Definition~\ref{def:SSNB}} We write a proof in the case where $\mathcal{E} = \{\Rd\}$. If $K > 1$, the proof can be applied independently on each set of the partition.

Let $(f_n)_{n \in \N}$ be such that $f_n(0) = 0$ for all $n \in \N$ and
\[
    \W_2\left[(\nabla f_n)_\sharp\mu, \nu\right] \leq \frac{1}{n+1} + \inf_{f \in \mathcal{F}_{\scvx, L}} \W_2\left[(\nabla f)_\sharp\mu, \nu\right].
\]
Let $x_0 \in \supp(\mu)$. Then there exists $C > 0$ such that for all $n \in \N$, $\|\nabla f_n(x_0)\| \leq C$. Indeed, suppose this is not true. Take $r > 0$ such that $V := \mu[B(x_0, r)] > 0$. By Prokhorov theorem, there exists $R > 0$ such that $\nu\left[B(0, R)\right] \geq 1-\frac{V}{2}$. Then for $C > 0$ large enough, there exists an $n \in \N$ such that:
\[
\begin{split}
    \W_2^2&[(\nabla f_n)_\sharp\mu, \nu] = \displaystyle\min_{\pi \in \Pi(\mu,\nu)} \int \|\nabla f_n(x) - y \|^2 \,d\pi(x,y) \\
    &\geq \displaystyle\int \|\nabla f_n(x) - \proj_{B(0, R)}\left[\nabla f_n(x)\right] \|^2 \,d\mu(x) \\
    &\geq \displaystyle\int_{B(x_0, r) \cap \supp(\mu)} \|\nabla f_n - \proj_{B(0, R)}\left[\nabla f_n\right] \|^2 \,d\mu \\
    &\geq \frac{1}{2} V \displaystyle\min_{\substack{x \in B(x_0, r)\\y \in B(0, R)}} \|\nabla f_n(x) - y \|^2 \\
    &\geq \frac{1}{2} V (C - Lr - R)
\end{split}
\]
which contradicts the definition of $f_n$ when $C$ is sufficiently large. \newline
Then for $x \in \Rd$,
\[
    \|\nabla f_n(x)\| \leq L \|x - x_0\| + \|\nabla f_n(x_0)\| \leq L \|x - x_0\| + C.
\]
Since $(\nabla f_n)_{n \in \N}$ is equi-Lipschitz, it converges uniformly (up to a subsequence) to some function $g$ by Arzel\`a–Ascoli theorem. Note that $g$ is $L$-Lipschitz.\newline
Let $\epsilon > 0$ and let $N \in \N$ such that $n \geq N \Rightarrow \|\nabla f_n - g\|_\infty \leq \epsilon$. Then for $n \geq N$ and $x \in \Rd$,
\[
    | f_n(x) | = \bigg| \int_0^1 \dotp{\nabla f_n(t x)}{x} \,dt \bigg| \leq \|x\| (\|g\|_\infty + \epsilon)
\]
so that $(f_n(x))$ converges up to a subsequence. Let $\phi, \psi$ be two extractions and $\alpha, \beta$ such that $f_{\phi(n)}(x) \to \alpha$ and $f_{\psi(n)}(x) \to \beta$. Then
\[
    \begin{split}
        | \alpha - \beta | &= \lim_{n \to \infty} \bigg| \int_0^1 \dotp{\nabla f_{\phi(n)}(t x) - \nabla f_{\psi(n)}(t x)}{x} \,dt \bigg| \\
        &\leq \lim_{n \to \infty} \|x\| \|\nabla f_{\phi(n)} - \nabla f_{\psi(n)}\|_\infty = 0.
    \end{split}
\]
This shows that $(f_n)_{n \in N}$ converges pointwise to some function $\opt{f}$. In particular, $\opt{f}$ is convex. For $x \in \Rd$, using Lebesgue's dominated convergence theorem,
\[
    \begin{split}
        \opt{f}(x) &= \lim_{n \to \infty} f_n(x) = \lim_{n \to \infty} \int_0^1 \dotp{\nabla f_n(t x)}{x} \,dt \\
        &= \int_0^1 \left\langle \lim_{n \to \infty}\nabla f_n(t x), x \right\rangle \,dt = \int_0^1 \dotp{g(t x)}{x} \,dt
    \end{split}
\]
so $\opt{f}$ is differentiable and $\nabla \opt{f} = g$. Using Lebesgue's dominated convergence theorem, uniform (hence pointwise) convergence of $(\nabla f_n)_{n \in \N}$ to $\nabla \opt{f}$ shows that $(\nabla f_n)_\sharp\mu \rightharpoonup (\nabla \opt{f})_\sharp\mu$. Then classical optimal transport stability theorems \emph{e.g.}~\citep[Theorem 5.19]{Villani09} show that
\[
    \begin{split}
        \W_2\left[(\nabla \opt{f})_\sharp\mu, \nu\right] &= \lim_{n \to \infty} \W_2\left[(\nabla f_n)_\sharp\mu, \nu\right]\\
        &= \inf_{f \in \mathcal{F}_{\scvx, L}} \W_2\left[(\nabla f)_\sharp\mu, \nu\right],
    \end{split}
\]
\emph{i.e.} $\opt{f}$ is a minimizer.

\paragraph{Proof of Theorem~\ref{thm:regasreg:optpb}} For $f \in \mathcal{F}_{\scvx, L,\mathcal{E}}$, $\nabla f_\sharp \mu = \sum_{i=1}^n a_i \delta_{\nabla f(x_i)}$. Writing $z_i = \nabla f(x_i)$, we wish to minimize $\W_2^2 \left( \sum_{i=1}^n a_i \delta_{z_i}, \nu \right)$ over all the points $z_1, \ldots, z_n \in \Rd$ such that there exists $f \in \mathcal{F}_{\scvx, L,\mathcal{E}}$ with $\nabla f(x_i) = z_i$ for all $i \in \range{n}$. Following~\citep[Theorem 3.8]{taylor2017convex}, there exists such a $f$ if, and only if, there exists $u \in \R^n$ such that for all $k \in \range{K}$ and for all $i,j \in I_k$,
\[
    \begin{split}
        u_i \geq & \text{ } u_j + \dotp{z_j}{x_i - x_j} + \frac{1}{2(1 - \scvx/L)} \left(
            \frac{1}{L} \|z_i - z_j\|^2 \right. \\
        &\left. + \scvx\|x_i - x_j\|^2
            - 2\frac{\scvx}{L} \dotp{z_j - z_i}{x_j - x_i} \right).
    \end{split}
\]
Then minimizing over $f \in \mathcal{F}_{\scvx, L, \mathcal{E}}$ is equivalent to minimizing over $(z_1, \ldots, z_n, u)$ under these interpolation constraints. \newline

The second part of the theorem is a direct application of~\citep[Theorem 3.14]{taylor2017convex}.

\paragraph{Proof of Proposition~\ref{prop:oneD:isotonic}} Let $f: \R\to\R$. Then $f \in \mathcal{F}_{\scvx, L,\mathcal{E}}$ if and only if it is convex and $L$-smooth on each set $E_k$, $k\in\range{K}$, \emph{i.e.} if and only if for any $k\in\range{K}$, $0 \leq \restriction{f''}{E_k} \leq L$.

For a measure $\rho$, let us write $F_\rho$ and $Q_\rho$ the cumulative distribution function and the quantile function (\emph{i.e.} the generalized inverse of the cumulative distribution function). Then $Q_{\nabla f_\sharp\mu} = \nabla f \circ Q_\mu$.

Using the closed-form formula for the Wasserstein distance in dimension $1$, the objective we wish to minimize (over $f \in \mathcal{F}_{\scvx, L,\mathcal{E}}$) is:
\[
    \W_2^2(f'_\sharp\mu, \nu) = \int_0^1 \left[ f' \circ Q_\mu(t) - Q_\nu(t) \right]^2 dt.
\]

Suppose $\mu$ has a density w.r.t the Lebesgue measure. Then by a change of variable,  the objective becomes
\[
    \int_{-\infty}^{+\infty} \left[ f'(x)  - Q_\nu \circ F_\mu (x) \right]^2 d\mu(x) = \| f' - \barycentric{\pi} \|^2_{L^2(\mu)}.
\]
Indeed, $Q_\nu \circ F_\mu$ is the optimal transport map from $\mu$ to $\nu$, hence its own barycentric projection. The result follows. \newline

Suppose now that $\mu$ is purely atomic, and write $\mu = \sum_{i=1}^n a_i \delta_{x_i}$ with $x_1 \leq \ldots \leq x_n$. For $0 \leq i \leq n$, put $\alpha_i = \sum_{k=1}^i a_k$ with $\alpha_0 = 0$. Then
\[
    \begin{split}
        \W_2^2(f'_\sharp\mu, \nu) &= \sum_{i=1}^{n} \int_{\alpha_{i-1}}^{\alpha_i} (f'(x_i) - Q_\nu(t))^2 dt\\
        &= \sum_{i=1}^{n} a_i \left[ f'(x_i) - \frac{1}{a_i} \left(\int_{\alpha_{i-1}}^{\alpha_i} Q_\nu(t) dt\right)\right]^2 \\
        &+ \int_{\alpha_{i-1}}^{\alpha_i} Q_\nu(t)^2 dt - \frac{1}{a_i} \left( \int_{\alpha_{i-1}}^{\alpha_i} Q_\nu(t) dt \right)^2.
    \end{split}
\]
Since $\sum_{i=1}^{n} \int_{\alpha_{i-1}}^{\alpha_i} Q_\nu(t)^2 dt - \frac{1}{a_i} \left( \int_{\alpha_{i-1}}^{\alpha_i} Q_\nu(t) dt \right)^2$ does not depend on $f$, minimizing $\W_2^2(f'_\sharp\mu, \nu)$ over $f \in \mathcal{F}_{\scvx, L,\mathcal{E}}$ is equivalent to solve 
\[
    \min_{f \in \mathcal{F}_{\scvx, L,\mathcal{E}}} \sum_{i=1}^{n} a_i \left[ f'(x_i) - \frac{1}{a_i} \left(\int_{\alpha_{i-1}}^{\alpha_i} Q_\nu(t) dt\right)\right]^2.
\]
There only remains to show that $\barycentric{\pi}(x_i) = \frac{1}{a_i} \int_{\alpha_{i-1}}^{\alpha_i} Q_\nu(t) \,dt$. Using the definition of the conditional expectation and the definition of $\pi$:
\begin{align*}
    \barycentric{\pi}(x_i) 
    &= \frac{1}{a_i} \int_{-\infty}^{+\infty} y\,\one{x=x_i} \,d\pi(x,y) \\
    &= \frac{1}{a_i} \int_{-\infty}^{+\infty} y\,\one{x=x_i} \,d(Q_\mu,Q_\nu)_\sharp\restriction{\Lebesgue{1}}{[0,1]} \\
    &= \frac{1}{a_i} \int_0^1 Q_\nu(t) \,\one{Q_\mu(t)=x_i} \,dt \\
    &= \frac{1}{a_i} \int_{\alpha_{i-1}}^{\alpha_i} Q_\nu(t) \,dt.
\end{align*}

\paragraph{Proof of Proposition~\ref{prop:estimation:L2}} Since $\mathcal{E} = \{\Rd\}$, and using the triangular inequality for the Wasserstein distance,
\begin{align}
    \left\vert \W_2(\mu, \nu) - \widehat \W_2(\mu,\nu) \right\vert
    &= \left\vert \W_2(\mu, \nu) - \W_2(\mu, {\nabla \hat f_n}_\sharp\mu) \right\vert \nonumber\\
    &\leq \W_2\left({\nabla \hat f_n}_\sharp\mu, \nu\right) \nonumber\\
    &\leq \W_2\left({\nabla \hat f_n}_\sharp\mu, {\nabla \hat f_n}_\sharp\hat\mu_n\right) \label{eqn:supplementary:prop_estimation:mu_mu_n}\\
    &\quad + \W_2\left({\nabla \hat f_n}_\sharp\hat\mu_n, \hat\nu_n\right) \label{eqn:supplementary:prop_estimation:mu_n_nu_n}\\
    &\quad + \W_2\left(\hat\nu_n, \nu\right). \label{eqn:supplementary:prop_estimation:nu_n_nu}
\end{align}

We now successively upper bound terms~\eqref{eqn:supplementary:prop_estimation:mu_mu_n},~\eqref{eqn:supplementary:prop_estimation:mu_n_nu_n},~\eqref{eqn:supplementary:prop_estimation:nu_n_nu}.

Since $\nabla \hat f_n$ is $L$-Lispchitz, almost surely:
\[
    \eqref{eqn:supplementary:prop_estimation:mu_mu_n}
    = \W_2\left({\nabla \hat f_n}_\sharp\mu, {\nabla \hat f_n}_\sharp\hat\mu_n\right)
    \leq L\,\W_2\left(\mu, \hat\mu_n \right)
    \underset{n \to \infty}{\longrightarrow} 0
\]
since almost surely, $\hat\mu_n \rightharpoonup \mu$ and $\mu$ has compact support, cf. \citep[Theorem 5.10]{SantambrogioBook}. For the same reason, almost surely:
\[
    \eqref{eqn:supplementary:prop_estimation:nu_n_nu} 
    = \W_2\left(\hat\nu_n, \nu\right)
    \underset{n \to \infty}{\longrightarrow} 0.
\]

Finally, since $\opt{f} \in \mathcal{F}_{\scvx, L, \mathcal{E}}$ and $\nabla \hat f_n$ is an optimal SSNB potential, it almost surely holds:
\begin{align*}
    \eqref{eqn:supplementary:prop_estimation:mu_n_nu_n}
    = \W_2\left({\nabla \hat f_n}_\sharp\hat\mu_n, \hat\nu_n\right)
    \leq \W_2\left({\nabla \opt{f}}_\sharp\hat\mu_n, \hat\nu_n\right) \\
    \underset{n \to \infty}{\longrightarrow} \W_2\left({\nabla \opt{f}}_\sharp\mu, \nu \right) = 0
\end{align*}
because $(\hat\mu_n, \hat\nu_n) \rightharpoonup (\mu, \nu)$, and by definition of $\opt{f}$, ${\nabla \opt{f}}_\sharp\mu = \nu$.

\end{document}